\documentclass[runningheads]{llncs}
\usepackage[T1]{fontenc}
\usepackage{subcaption}
\usepackage{float}

\usepackage{graphicx}
\usepackage{orcidlink}
\usepackage{amsfonts,amsmath,amssymb}
\usepackage{bbm}
\usepackage{tabularx}
\usepackage{hyperref}
\usepackage{makecell}
\usepackage{booktabs,array}
\usepackage{textcomp}
\usepackage{color}

\begin{document}
\title{Towards Robust Cross-Dataset Object Detection Generalization under Domain Specificity}
\titlerunning{Cross-Domain Object Detection}

\author{
    Ritabrata Chakraborty\inst{1,2}\orcidlink{0009-0009-3597-3703}\thanks{ denotes equal contribution} \and
    Hrishit Mitra\inst{4}\orcidlink{0009-0009-3597-3703}*\and
 Shivakumara Palaiahnakote\inst{3}\orcidlink{0000-0001-9026-4613} \and
    Umapada Pal\inst{1}\orcidlink{0000-0002-5426-2618}
}

\institute{
    CVPR Unit, Indian Statistical Institute, Kolkata, India \\
    \and
    Manipal University Jaipur, India \\
    \and
    University of Salford, UK \\
    \and
    National Institute of Technology, Trichy\\
    \vspace{0.3cm}
\email{ritabrata.229301716@muj.manipal.edu}
}
\authorrunning{R. Chakraborty et al.}
%
\maketitle              
\setcounter{footnote}{0}  
%
\begin{abstract}

Object detectors often perform well in-distribution, yet degrade sharply on a different benchmark. We study cross-dataset object detection (CD-OD) through a lens of setting specificity. We group benchmarks into setting-agnostic datasets with diverse everyday scenes and setting-specific datasets tied to a narrow environment, and evaluate a standard detector family across all train--test pairs. This reveals a clear structure in CD-OD: transfer within the same setting type is relatively stable, while transfer across setting types drops substantially and is often asymmetric. The most severe breakdowns occur when transferring from specific sources to agnostic targets, and persist after open-label alignment, indicating that domain shift dominates in the hardest regimes. To disentangle domain shift from label mismatch, we compare closed-label transfer with an open-label protocol that maps predicted classes to the nearest target label using CLIP similarity. Open-label evaluation yields consistent but bounded gains, and many corrected cases correspond to semantic near-misses supported by the image evidence. Overall, we provide a principled characterization of CD-OD under setting specificity and practical guidance for evaluating detectors under distribution shift. Code will be released at \href{[https://github.com/Ritabrata04/cdod-icpr.git}{https://github.com/Ritabrata04/cdod-icpr}

\end{abstract}

%
%
\section{Introduction}\label{sec:intro}

\begin{table}[t]
\centering
\scriptsize
\setlength{\tabcolsep}{4.5pt}
\caption{Human study summary. }
\label{tab:human_study_summary}

\begin{minipage}[t]{0.48\linewidth}
\centering
\textbf{Likert responses} \\[2pt]
\begin{tabular}{lc}
\toprule
\textbf{Metric} & \textbf{Value} \\
\midrule
Mean & 4.83 \\
Median [IQR] & 4 [2, 8] \\
Accuracy (exact GT) & 35.3\% \\
Agreement (within $\pm$1 of GT) & 65.4\% \\
\bottomrule
\end{tabular}
\end{minipage}
\hfill
\begin{minipage}[t]{0.48\linewidth}
\centering
\textbf{Yes/No responses} \\[2pt]
\begin{tabular}{lc}
\toprule
\textbf{Metric} & \textbf{Value} \\
\midrule
Accuracy & 84.0\% \\
Accuracy (GT = Yes) & 80.3\% \\
Accuracy (GT = No) & 89.7\% \\
Per-question accuracy range & 66.2--98.7\% \\
\bottomrule
\end{tabular}
\end{minipage}

\end{table}

Object detection is a safety critical capability in deployed perception systems, and when it fails under rare conditions the consequences can be immediate. In late October 2025, public reporting and released surveillance footage describe a driverless Waymo vehicle in San Francisco fatally injuring a well known neighborhood cat, KitKat, after the animal moved beneath the vehicle while it was stopped and the vehicle then pulled away.\footnote{\url{https://www.nytimes.com/2025/12/05/us/waymo-kit-kat-san-francisco.html}}
This incident is not a basis for attributing failure to any single component, but it highlights a broader deployment reality in which real scenes contain unusual occlusions, rare object trajectories, and operating conditions that are weakly represented in common evaluation suites.

Despite this gap, modern detectors are typically developed and benchmarked within a single dataset ecosystem, which has produced steady progress on standard leaderboards while leaving an important question unresolved. \textbf{How does a detector trained on one benchmark behave when evaluated on a different benchmark that reflects different environments and visual statistics while also using a different category vocabulary?}
Cross dataset results are also difficult to interpret because two effects are entangled. Performance can drop due to visual distribution shift, but it can also drop due to label taxonomy shift since datasets frequently name, split, or merge concepts differently, so a detector can output a semantically plausible label and still be scored as incorrect under a strict target vocabulary. Transfer tables alone therefore conflate genuine appearance mismatch with evaluation artifacts.

\textbf{We propose \textbf{setting specificity} as a lens that makes cross dataset results easier to explain, defining it through the intention of the dataset collection.} A dataset is setting specific when it targets a core operational setting with repeated scene structure, consistent viewpoints, and a task focused taxonomy, whereas a dataset is setting agnostic when it is collected to cover many settings so capture conditions vary widely and the label space is broader and less concentrated.

This distinction is perceptually salient, which we verify with a human study of 78 participants (shown in Table \ref{tab:human_study_summary}) using curated images from COCO \cite{lin2014microsoft}, Cityscapes \cite{cordts2015cityscapes}, Objects365 \cite{shao2019objects365}, and BDD100K \cite{yu2018bdd100k}. Participants rated visual diversity and judged whether image pairs came from the same dataset, and we observe strongly polarized Likert ratings across sets with high agreement around the per question mode as well as high agreement with the per question majority label in the Yes or No task, with higher agreement for different dataset pairs than for same dataset pairs. These results support our hypothesis that datasets exhibit stable setting level signatures that humans can recover from visual cues alone, motivating the question of whether detector generalization is similarly organized by setting specificity. We test this hypothesis with a compact transfer grid that covers all combinations of setting types, including within type transfer where capture geography, weather, sensors, and annotation policy can still shift. 

Overall, our contributions are as follows:
\begin{itemize}
\item \textbf{A setting-aware view of cross-dataset detection.} We formalize setting specificity as a dataset-level factor and use it to structure cross-dataset evaluation beyond dataset identity.
\item \textbf{Consistent transfer organization by setting type.} We find stable train to test transfer patterns across all setting-type pairings, including informative within-type degradations.
\item \textbf{Separating taxonomy effects from visual generalization.} We pair closed label evaluation with an open label protocol based on semantic label similarity to estimate how much transfer loss is driven by label mismatch rather than appearance shift.
\end{itemize}

The rest of the paper is structured as follows. Section \ref{sec:related_work} describes the related literature for our work. Section~\ref{sec: experimental_setup} describes our setup, including the task formulation, setting specificity definition, datasets, training protocol, and evaluation metrics. Section~\ref{sec:quant} presents quantitative transfer results and ablations. Section~\ref{sec:qual} provides a qualitative analysis. Section~\ref{sec: discussion} discusses implications for the community, and Section~\ref{sec:conclusion} concludes the paper.

\begin{figure}[t]
    \centering

    \begin{subfigure}{0.23\linewidth}
        \centering
        \includegraphics[width=\linewidth]{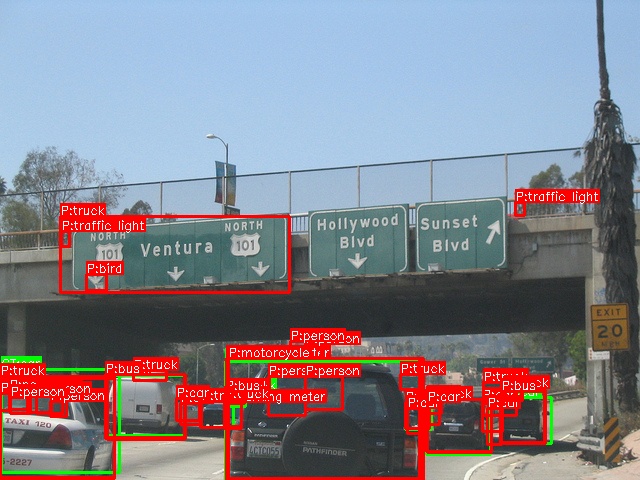}
        \caption{}
    \end{subfigure}\hfill
    \begin{subfigure}{0.23\linewidth}
        \centering
        \includegraphics[width=\linewidth]{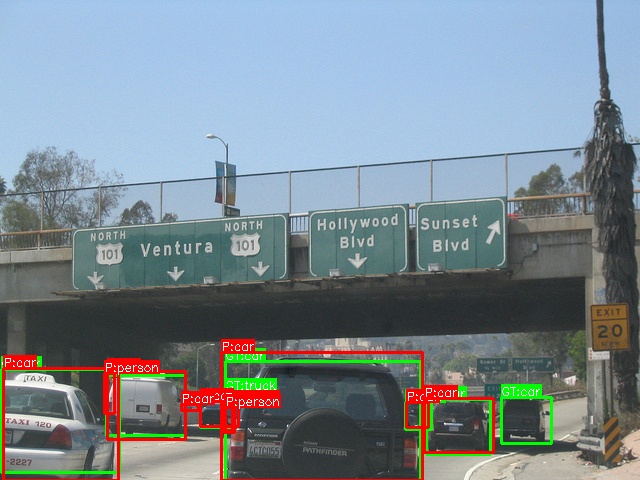}
        \caption{}
    \end{subfigure}\hfill
    \begin{subfigure}{0.23\linewidth}
        \centering
        \includegraphics[width=\linewidth]{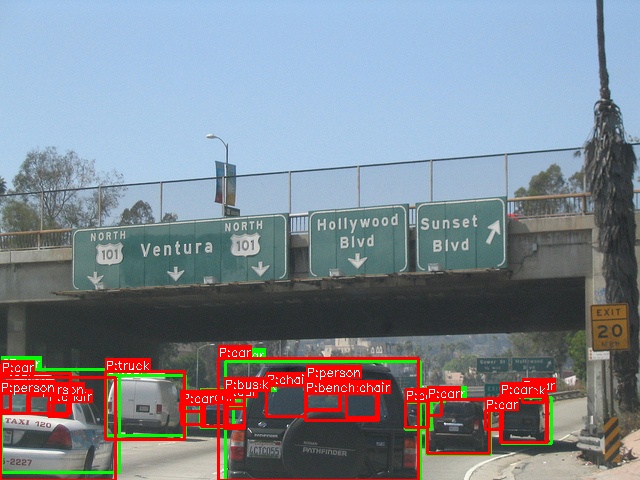}
        \caption{}
    \end{subfigure}\hfill
    \begin{subfigure}{0.23\linewidth}
        \centering
        \includegraphics[width=\linewidth]{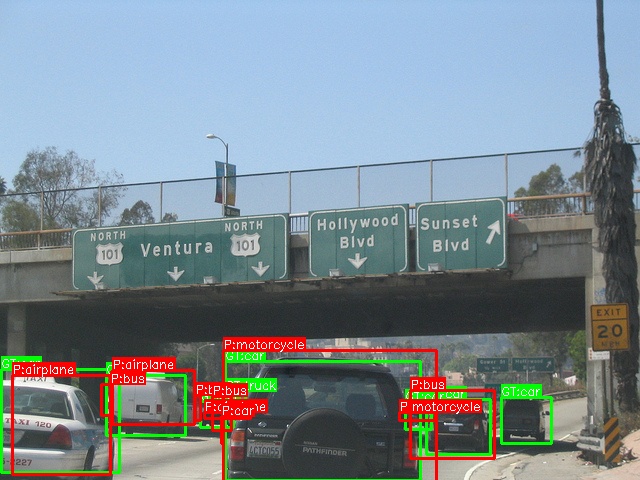}
        \caption{}
    \end{subfigure}

    \caption{Object detector inference on an image from the COCO dataset. (a) shows the In-domain performance of COCO, whereas (b), (c), and (d) show the Cross-domain performance for a model trained on Cityscapes, Objects365, and BDD100k respectively.}

\end{figure}

\section{Related Work}
\label{sec:related_work}

\subsection{Robustness and Dataset Bias}

A growing body of work has shown that strong in-distribution performance does
not translate to robustness under naturally occurring distribution shifts.
Early studies on dataset bias demonstrated that recognition and detection
models often exploit dataset-specific correlations rather than transferable
semantics, leading to severe generalization failures across datasets.
~\cite{torralba2011unbiased} formalized this issue by showing that cross-dataset evaluation reveals substantial bias even when label spaces are aligned.

More recently, robustness benchmarks such as WILDS~\cite{koh2021wilds} and COCO-O~\cite{mao2023coco} systematize evaluation under real-world distribution shifts. WILDS covers shifts across time, geography, and data sources, while COCO-O probes appearance and context changes without synthetic perturbations~\cite{mao2023coco}.

However, these investigations typically evaluate under controlled shift
circumstances, often within a single dataset family. They do not directly
address the task of cross-dataset object detection, where models trained on
one dataset must generalize to entirely different data collection pipelines,
which is more aligned with real-world deployment scenarios.

\subsection{Domain Adaptation for Object Detection}

Domain adaptation (DA) addresses distribution shift by assuming access to
unlabeled or weakly labeled target data. Foundational adversarial alignment
methods laid the groundwork for detector-specific adaptations, including Domain Adaptive Faster R-CNN~\cite{chen2018domain},
strong-weak alignment~\cite{saito2019strong},
selective alignment~\cite{zhu2019adapting},
categorical regularization~\cite{xu2020exploring},
progressive adaptation~\cite{hsu2020progressive},
curriculum learning~\cite{soviany2021curriculum},
teacher-student frameworks~\cite{li2022cross},
self-training~\cite{yu2019unsupervised},
robust DA~\cite{khodabandeh2019robust},
source-free DA~\cite{li2021free},
transformer-based DA~\cite{jia2023pm},
pixel-level alignment~\cite{hsu2020every},
spatial attention~\cite{Li2020SpatialAPY},
and adverse-weather adaptation~\cite{Sindagi2019PriorBasedDAZ}.

While DA methods demonstrate impressive gains on standard benchmarks such as
Sim10k~\cite{tsai2018learning}$\rightarrow$Cityscapes~\cite{cordts2015cityscapes} or Syn2Real~\cite{peng2018syn2real}, they fundamentally rely on
target-domain access, either offline or online. As a result, their reported
performance conflates robustness with adaptation, making it difficult to
assess how detectors behave when deployed on unseen domains without prior
exposure. Our benchmark explicitly isolates this missing evaluation regime.

\subsection{Domain Generalization and Test-Time Adaptation}

Domain generalization (DG) assumes no target-domain access and learns representations invariant across multiple source domains. Detection DG spans feature diversification, alignment regularization, transformer-based designs, and diffusion-guided representations, with semi-supervised variants adding auxiliary cues such as language~\cite{mohamadi2024feature}. Test-time adaptation (TTA) allows limited model updates at inference, but can incur error accumulation and deployment instability, especially for detection. Most DG and TTA evaluations rely on predefined benchmark splits that may understate real cross-dataset shift. Our protocol complements this work by testing frozen detectors across heterogeneous datasets, exposing failure modes that adaptation-based evaluations can hide.




\subsection{Cross-Domain and Cross-Dataset Object Detection}

Cross-domain object detection transfers detectors across datasets with minimal or no target supervision. Benchmarks such as Syn2Real and cross-domain document detection show that domain gaps can be severe even within a single application vertical, motivating alignment, distillation, and disentanglement strategies~\cite{ramamonjison2021simrod,jiang2021decoupled,Krishna2025VisuallySPM,Shi2025EmbodiedDAO,Lu2023FedDADFDQ,Hu2023DensityInsensitiveUDK,Chen2023RevisitingD3L,Zhang2024DetectCSP}. Despite this progress, evaluations often rely on a small set of curated source--target pairs and aggregate metrics that mask cross-domain instability, and many protocols implicitly allow some form of adaptation.

In contrast, our benchmark isolates cross-dataset robustness under a unified training-free protocol by evaluating frozen detectors across diverse datasets without target access, showing that modern methods remain brittle in the wild despite advances in domain generalization, adaptation, and open-vocabulary learning.

\section{Methodological Setup}
\label{sec: experimental_setup}

\subsection{Task Definition and Setting Specificity}
\label{sec:task_setting}

Let $\mathcal{D}_s$ and $\mathcal{D}_t$ denote a source and target detection dataset with image spaces $\mathcal{X}_s,\mathcal{X}_t$ and label vocabularies $\mathcal{Y}_s,\mathcal{Y}_t$. In cross-dataset object detection, we train a detector on $\mathcal{D}_s$ and evaluate on $\mathcal{D}_t$, measuring transfer performance under a target evaluation protocol. We report results for all ordered pairs $(\mathcal{D}_s,\mathcal{D}_t)$ in a fixed transfer grid.

We organize datasets by \emph{setting specificity}, defined by the intention of dataset collection. A dataset is \textit{setting specific} if it targets a single operational setting with repeated scene structure, consistent viewpoints, and a task focused taxonomy. A dataset is \textit{setting agnostic} if it is collected to span many settings and scene types, producing broader variation in capture conditions and a less concentrated label space. This induces four transfer regimes given a source type $S\in\{\text{specific},\text{agnostic}\}$ and target type $T\in\{\text{specific},\text{agnostic}\}$, namely $S\!\rightarrow\!T$.

\begin{table}[ht]
\centering
\small
\setlength{\tabcolsep}{6pt}
\begin{tabular}{lrrc}
\toprule
\textbf{Dataset} & \textbf{Images} & \textbf{Classes} & \textbf{Setting} \\
\midrule
COCO              & 5{,}000  & 80  & Agnostic \\
Objects365        & 38{,}000 & 365 & Agnostic \\
\midrule
Cityscapes        & 500      & 8   & Specific \\
BDD100k           & 10{,}000 & 10  & Specific \\
\bottomrule
\end{tabular}
\caption{Dataset statistics for evaluation datasets used in our experiments.
\textit{Setting} distinguishes setting-agnostic datasets with generic objects from domain-specific datasets such as driving. The 'Classes' here refer to only the instance-level classes in each dataset.}
\label{tab:dataset-stats}
\end{table}

\begin{figure}[ht]
    \centering

    \begin{subfigure}{0.23\linewidth}
        \centering
        \includegraphics[width=\linewidth]{b1d0a191-65deaeef.jpg}
        \caption{}
    \end{subfigure}\hfill
    \begin{subfigure}{0.23\linewidth}
        \centering
        \includegraphics[width=\linewidth]{b63368bc-37bc42c5.jpg}
        \caption{}
    \end{subfigure}\hfill
    \begin{subfigure}{0.23\linewidth}
        \centering
        \includegraphics[width=\linewidth]{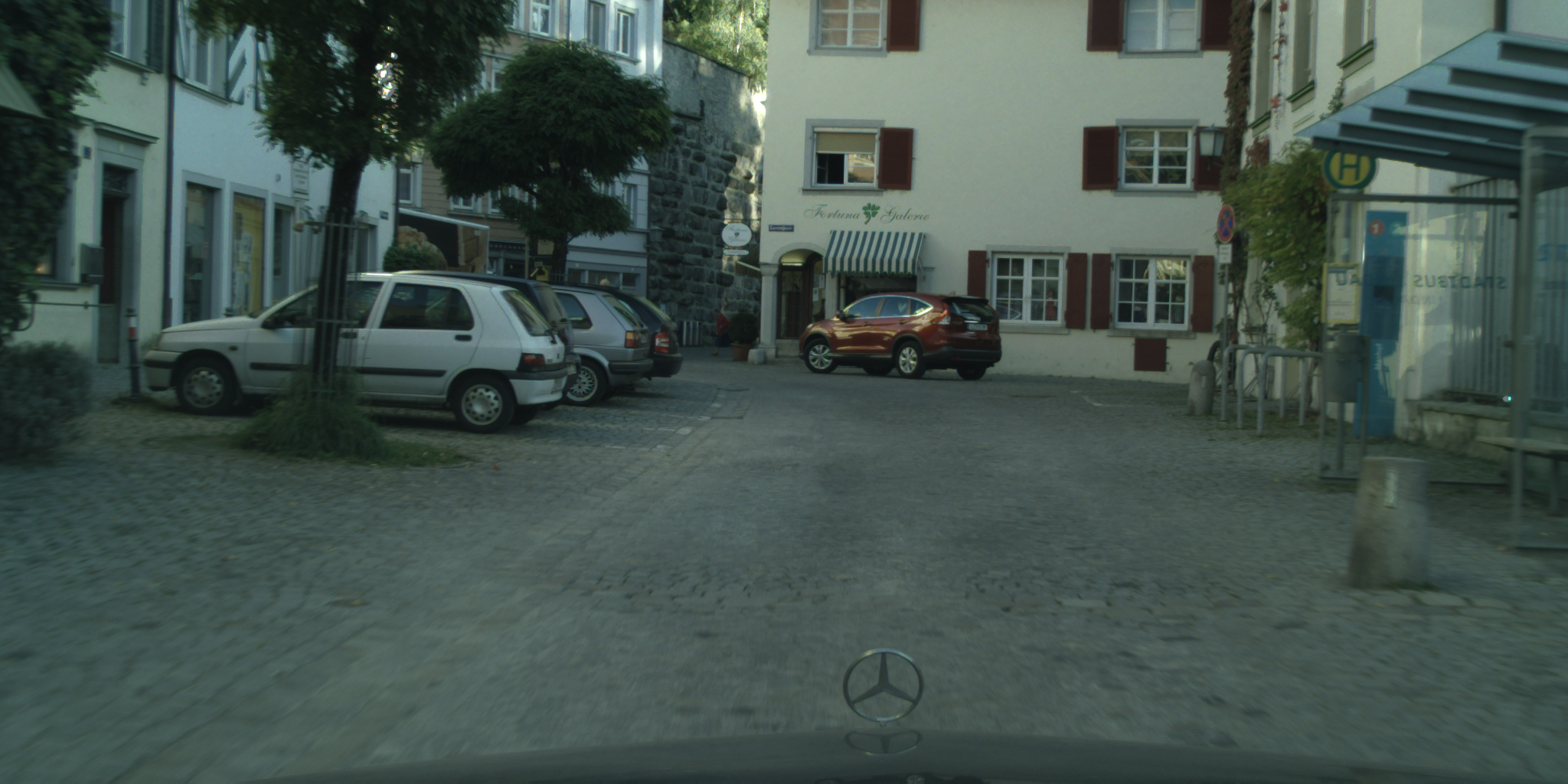}
        \caption{}
    \end{subfigure}\hfill
    \begin{subfigure}{0.23\linewidth}
        \centering
        \includegraphics[width=\linewidth]{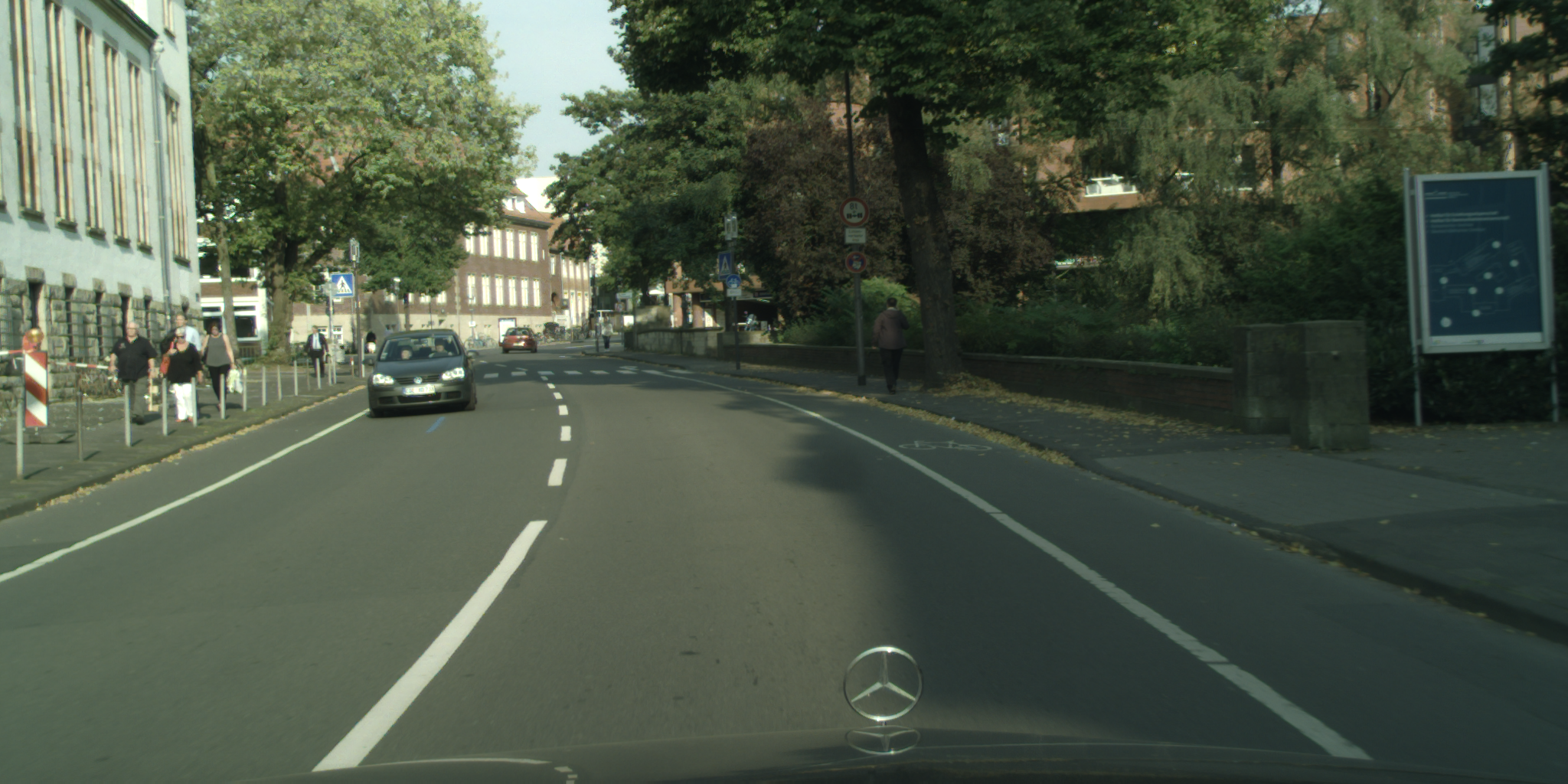}
        \caption{}
    \end{subfigure}

    \caption{Setting-specific examples across different domains. (a) and (b) are BDD100k images, whereas (c) and (d) are Cityscapes images}
    \label{fig: specific}

\end{figure}

\begin{figure}[t]
    \centering

    \begin{subfigure}{0.23\linewidth}
        \centering
        \includegraphics[width=\linewidth]{000000019742.jpg}
        \caption{}
    \end{subfigure}\hfill
    \begin{subfigure}{0.23\linewidth}
        \centering
        \includegraphics[width=\linewidth]{000000441247.jpg}
        \caption{}
    \end{subfigure}\hfill
    \begin{subfigure}{0.23\linewidth}
        \centering
        \includegraphics[width=\linewidth]{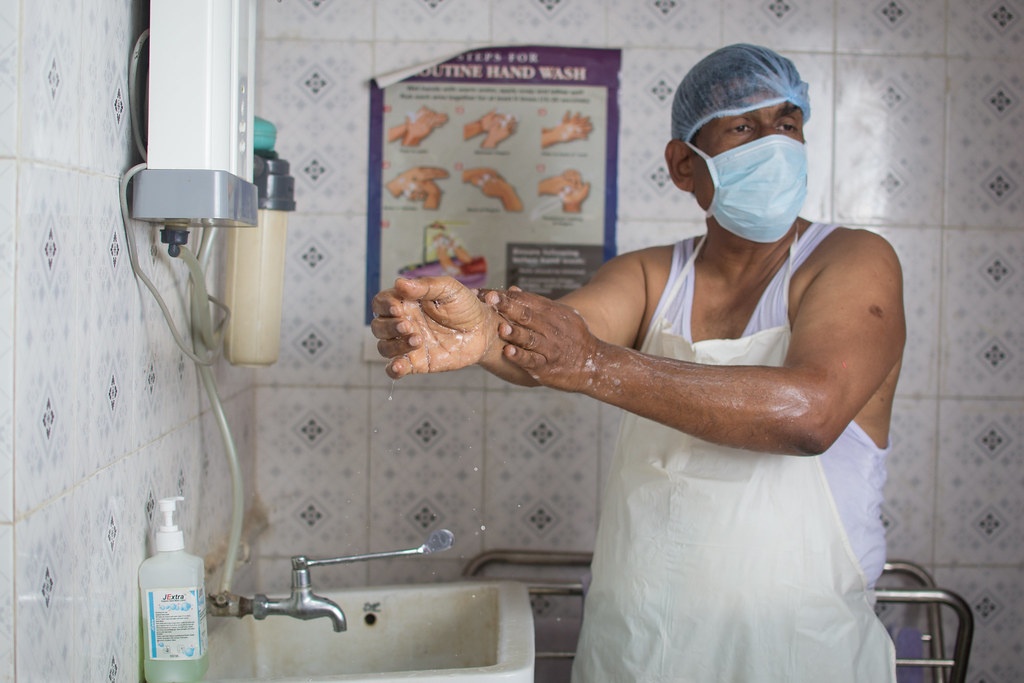}
        \caption{}
    \end{subfigure}\hfill
    \begin{subfigure}{0.23\linewidth}
        \centering
        \includegraphics[width=\linewidth]{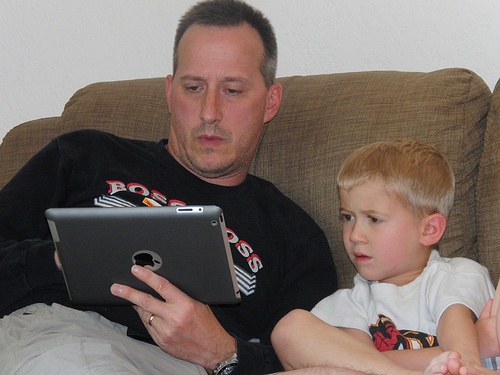}
        \caption{}
    \end{subfigure}\hfill

    \caption{Setting-agnostic examples. (a) and (b) are COCO images whereas (c) and (d) are Objects365 images}
    \label{fig: agnostic}

\end{figure}

To study cross-domain generalisation under varying degrees of semantic and contextual diversity, we employ two domain-specific datasets with limited ontologies, Cityscapes and BDD100k, and two general-purpose object detection datasets, COCO and Objects365. This selection allows us to systematically contrast model behaviour across datasets that differ along axes of domain-focus, ontology size, contextual priors, and long-tail object diversity.

\subsection{Datasets}
\label{sec:data}
We evaluate four widely used benchmarks that cover both ends of the setting specificity spectrum. COCO and Objects365 represent setting-agnostic collections with broad object coverage and diverse contexts. Cityscapes and BDD100k represent setting-specific driving collections with structured camera viewpoints and a driving-centric taxonomy.

Across all dataset pairs, we treat taxonomy differences as a first-class confounder during evaluation rather than silently merging concepts. As illustrated in Table \ref{tab:dataset-stats}, COCO and Objects365 (Figure \ref{fig: agnostic}) can both be classified as setting-agnostic datasets based on the size and diversity of their label spaces. However, Objects365 significantly expands the object vocabulary relative to COCO, particularly in terms of fine-grained and rare categories, and encompasses objects drawn from a wide range of functional and environmental contexts. 
Cityscapes and BDD100k (Figure \ref{fig: specific})  both have limited and strongly overlapping label-spaces. Cityscapes provides instance-level annotations for only eight object categories, which are all fully contained within the BDD100k label space, which additionally includes only two more labels. All instance-level categories in both datasets are explicitly driving-centric, resulting in a limited, domain-focused ontology.
Despite their similarity in label spaces, BDD100k possesses significantly greater diversity in capture conditions when compared with Cityscapes. BDD100k includes both daytime and nighttime scenes, as well as adverse weather conditions such as rain and snow, enabling evaluation under challenging visual conditions.

\subsection{Training protocol}
\label{sec:train}
To isolate dataset effects, we keep the training recipe fixed across all source datasets and train one model per source benchmark. We use a uniform training schedule and run all training and inference through a single codebase to avoid toolkit-dependent variation. All models are trained only on the source dataset, and all target evaluations are zero-shot with no target access. Faster R-CNN with a ResNet50 FPN backbone \cite{ren2016fasterrcnnrealtimeobject} was used as a common architecture due to its stability and widespread adoption in detection benchmarks, its explicit decoupling of localization and classification, and its balanced model capacity that enables fair and controlled comparison across datasets without confounding architectural effects.

\subsection{Implementation Details}
\label{sec:imp}
All experiments are implemented in PyTorch using the MMDetection framework with a Faster R-CNN ResNet-50 FPN backbone and a standard 1× learning-rate schedule. Evaluations are performed on the validation splits of COCO, Cityscapes, Objects365v2, and BDD100K. In ablations, we compute semantic similarity using CLIP ViT-L/14\cite{radford2021learning} text embeddings with raw label strings as prompts.
We apply open-label remapping after NMS, leaving all boxes and IoU matching unchanged.Experiments are conducted on an NVIDIA RTX A5000 GPU (24 GB VRAM).

\subsection{Evaluation protocols and metrics}
\label{sec:eval}
We report a transfer grid over all train to test pairs, including transfers within the same setting type and across setting types.

\paragraph{Closed-label evaluation.}
We follow COCO-style detection metrics and report AP averaged over IoU thresholds from 0.50 to 0.95 and size-specific AP for small, medium, and large objects. Since datasets differ in taxonomies, closed-label evaluation restricts scoring to the one-to-one intersection of shared labels between source and target. We implement label alignment using explicit mapping files with case-insensitive normalization, we do not perform semantic collapsing, and we filter out predictions whose labels lie outside the shared set before scoring.

\paragraph{Open-label evaluation.}
Closed-label transfer can penalize semantically plausible predictions that fail exact label matching. To estimate how much cross-dataset loss is driven by taxonomy mismatch, we also report an open-label protocol that re-aligns predicted classes to the nearest target label using semantic similarity between label texts. Concretely, for a predicted label $\hat{y}$ we compute similarity to each target label $c \in \mathcal{Y}_t$ and assign $\hat{y}$ to the nearest target label when the best similarity exceeds a threshold $\tau$. This protocol changes label scoring while leaving localization unchanged.

\paragraph{Semantic near-miss diagnostics.}
To validate that open-label gains correspond to meaningful near-misses rather than arbitrary remapping, we summarize label-mismatch cases with two CLIP-based diagnostics. We compute text-to-text similarity
\[
s_{tt}(\hat{y},y)=\cos(e_T(\hat{y}), e_T(y)),
\]
the rank of the ground-truth label $y$ under the predicted label $\hat{y}$,
\[
\text{rank}_{\mathrm{GT}\mid \hat{y}} = 1 + \sum_{c \in \mathcal{Y}_t} \mathbbm{1}\!\left[\cos(e_T(\hat{y}),e_T(c))>\cos(e_T(\hat{y}),e_T(y))\right],
\]
and a region-level preference margin on the predicted region crop $r$,
\[
s_{it}(r,c)=\cos(e_I(r),e_T(c)), \qquad \Delta_{it}=s_{it}(r,y)-s_{it}(r,\hat{y}).
\]
A mismatch is treated as a semantic near-miss when $s_{tt}\ge \tau$ and $\text{rank}_{\mathrm{GT}\mid \hat{y}}\le K$, and we report the near-miss rate along with summary statistics of $s_{tt}$, rank, and $\Delta_{it}$.

\section{Quantitative Results}
\label{sec:quant}



\begin{table}[t]
\centering
\small
\setlength{\tabcolsep}{6pt}
\caption{Cross-dataset object detection performance (mAP @ 0.50--0.95) on closed-label evluation setting.}
\label{tab:cross_dataset}
\begin{tabular}{llccccc}
\toprule
 &  & \multicolumn{4}{c}{\textbf{Train}} &  \\
\cmidrule(lr){3-6}
\textbf{Test} &  & COCO\cite{lin2014microsoft} & Cityscapes\cite{cordts2015cityscapes} & Objects365\cite{shao2019objects365} & BDD\cite{yu2018bdd100k} & \textbf{Avg.} \\
\midrule
COCO\cite{lin2014microsoft}        &  & 0.376 & 0.015 & 0.286 & 0.019 & 0.174 \\
Cityscapes\cite{cordts2015cityscapes}   &  & 0.206 & 0.402 & 0.268 & 0.275 & 0.287 \\
Objects365\cite{shao2019objects365} &  & 0.046 & 0.020 & 0.198 & 0.102 & 0.092 \\
BDD\cite{yu2018bdd100k}          &  & 0.131 & 0.025 & 0.103 & 0.298 & 0.139 \\
\midrule
\textbf{Avg.} &  & 0.189 & 0.115 & 0.213 & 0.174 & -- \\
\bottomrule
\end{tabular}
\end{table}

\begin{table}[ht]
\centering
\small
\setlength{\tabcolsep}{6pt}
\caption{Cross-dataset object detection performance (mAP\@0.50--0.95) under \textbf{open-label} evaluation (cosine similarity $\ge 0.6$ to nearest test label). Diagonal entries are unchanged. Red subscripts denote gains over closed-label evaluation.}
\label{tab:cross_dataset_openlabel}
\begin{tabular}{llccccc}
\toprule
 &  & \multicolumn{4}{c}{\textbf{Train}} &  \\
\cmidrule(lr){3-6}
\textbf{Test} &  & COCO\cite{lin2014microsoft} & Cityscapes\cite{cordts2015cityscapes} & Objects365\cite{shao2019objects365} & BDD\cite{yu2018bdd100k} & \textbf{Avg.} \\
\midrule
COCO\cite{lin2014microsoft}         &  & \textcolor{gray}{0.376} & 0.030$_{\color{red}\scriptstyle +0.015}$ & 0.322$_{\color{red}\scriptstyle +0.036}$ & 0.038$_{\color{red}\scriptstyle +0.019}$ & 0.192 \\
Cityscapes\cite{cordts2015cityscapes}   &  & 0.235$_{\color{red}\scriptstyle +0.029}$ & \textcolor{gray}{0.402} & 0.296$_{\color{red}\scriptstyle +0.028}$ & 0.287$_{\color{red}\scriptstyle +0.012}$ & 0.305 \\
Objects365\cite{shao2019objects365}  &  & 0.082$_{\color{red}\scriptstyle +0.036}$ & 0.032$_{\color{red}\scriptstyle +0.012}$ & \textcolor{gray}{0.198} & 0.124$_{\color{red}\scriptstyle +0.022}$ & 0.109 \\
BDD\cite{yu2018bdd100k}           &  & 0.160$_{\color{red}\scriptstyle +0.029}$ & 0.040$_{\color{red}\scriptstyle +0.015}$ & 0.132$_{\color{red}\scriptstyle +0.029}$ & \textcolor{gray}{0.298} & 0.158 \\
\midrule
\textbf{Avg.} &  & 0.213 & 0.126 & 0.237 & 0.187 & -- \\
\bottomrule
\end{tabular}
\end{table}

\subsection{Cross-Dataset Generalization Analysis}
\begin{table}[t]
\centering
\small
\setlength{\tabcolsep}{3pt}
\renewcommand{\arraystretch}{0.95}
\caption{Open-label diagnostics using CLIP. We analyze matched detections where the box is correct (IoU matched) but the label may differ. Near-miss rate is the fraction of label mismatches that satisfy $s_{tt}\ge 0.6$ and $\mathrm{rank}_{GT|\hat{y}}\le 5$.}
\label{tab:clip_ablation}
\begin{tabularx}{\linewidth}{l c c c c c >{\raggedright\arraybackslash}X}
\toprule
\textbf{Train$\rightarrow$Test} &
\makecell{$\Delta$mAP} &
\makecell{\textbf{Near-miss}\\$\uparrow$} &
\makecell{$\mathbb{E}[s_{tt}]$\\$\uparrow$} &
\makecell{Med.\ Rank\\$_{GT|\hat{y}}$ $\downarrow$} &
\makecell{Top-5\\$\uparrow$} &
\makecell{$\mathbb{E}[\Delta_{it}]$\\(\%($\Delta_{it}>0$))} \\
\midrule
Objects365 $\rightarrow$ BDD   & \textcolor{red}{+0.25} & 41\% & 0.73 & 2  & 88\% & +0.018 (62\%) \\
Objects365 $\rightarrow$ City  & \textcolor{red}{+0.15} & 33\% & 0.71 & 3  & 81\% & +0.012 (58\%) \\
COCO $\rightarrow$ Objects365  & \textcolor{red}{+0.04} & 18\% & 0.69 & 4  & 64\% & +0.006 (55\%) \\
Objects365 $\rightarrow$ COCO  & \textcolor{red}{+0.04} & 17\% & 0.70 & 4  & 66\% & +0.005 (54\%) \\
City $\rightarrow$ COCO        & \textcolor{red}{+0.01} &  4\% & 0.58 & 14 & 21\% & -0.002 (46\%) \\
\bottomrule
\end{tabularx}
\end{table}

\begin{table}[ht]
\centering
\small
\setlength{\tabcolsep}{6pt}
\caption{Sensitivity to semantic threshold $\tau$ for open-label matching (cosine similarity score on CLIP text embeddings).}
\label{tab:tau_sweep}
\begin{tabular}{lcccc}
\toprule
\textbf{Pair} & $\tau=0.5$ & $\tau=0.6$ & $\tau=0.7$ & \textbf{Closed} \\
\midrule
Objects365$\rightarrow$BDD  & 0.565 & \textbf{0.580} & 0.522 & 0.298 \\
Objects365$\rightarrow$City & 0.430 & \textbf{0.475} & 0.398 & 0.268 \\
COCO$\rightarrow$O365       & \textbf{0.243} & 0.242 & 0.226 & 0.198 \\
\bottomrule
\end{tabular}
\end{table}

Table~\ref{tab:cross_dataset} reports closed-label cross-dataset transfer, where evaluation is restricted to the shared label intersection for each train--test pair. Even under this conservative protocol, cross-dataset performance drops sharply relative to in-domain scores, indicating that the dominant failure mode is not only taxonomy mismatch but also appearance and capture bias.

Across sources, Cityscapes is the most transferable training set in this grid, achieving the highest average performance (0.287) while retaining strong transfer to both driving and general-purpose targets. In particular, Cityscapes$\rightarrow$BDD reaches 0.275 and Cityscapes$\rightarrow$Objects365 reaches 0.268, both substantially higher than transfers originating from the general-purpose sources. In contrast, Objects365 is the weakest source overall, with very low transfer to COCO (0.046) and Cityscapes (0.020), despite a reasonable in-domain score of 0.198. Transfer is also strongly asymmetric, including within the same setting type. For the agnostic pair, COCO$\rightarrow$Objects365 (0.286) is far stronger than Objects365$\rightarrow$COCO (0.046). For the specific pair, Cityscapes$\rightarrow$BDD (0.275) is far stronger than BDD$\rightarrow$Cityscapes (0.025). This asymmetry suggests that cross-dataset generalization depends on more than setting identity, and is sensitive to annotation policy, object scale distributions, and the dataset-specific priors a detector learns during training. Finally, target difficulty varies substantially. Averaged over all sources, Cityscapes is the hardest target in this grid (0.115), while Objects365 is the easiest (0.213). This pattern is consistent with the hypothesis that narrow, viewpoint-constrained driving benchmarks induce stronger dataset signatures that do not readily align with learned representations from other collections.

\subsection{Open-Label Cross-Dataset Evaluation}
\label{sec:open_label_results}

Closed-label transfer penalizes semantically plausible predictions that fail exact name matching across taxonomies. To estimate how much transfer loss is attributable to this effect, Table~\ref{tab:cross_dataset_openlabel} reports open-label evaluation, where predicted class names are mapped to the nearest target label by CLIP text similarity (with $\tau=0.6$), leaving localization unchanged. Open-label matching yields consistent but bounded gains on all off-diagonal entries. The improvements are largest for transfers between the two agnostic datasets and for transfers from driving datasets into the broader vocabularies. For example, COCO$\rightarrow$Objects365 improves from 0.286 to 0.322 (+0.036), and Objects365$\rightarrow$COCO improves from 0.046 to 0.082 (+0.036). Cityscapes$\rightarrow$COCO improves from 0.206 to 0.235 (+0.029), while BDD$\rightarrow$COCO improves from 0.131 to 0.160 (+0.029). These gains indicate that a meaningful portion of cross-dataset error is due to label mismatch and naming granularity rather than incorrect boxes.

At the same time, open-label evaluation does not close the robustness gap. Several transfers remain extremely low even after semantic remapping, such as COCO$\rightarrow$Cityscapes (0.030) and COCO$\rightarrow$BDD (0.038). This residual gap points to harder transfer phenomena, including viewpoint and context shift, long-tail frequency changes, and localization degradation that cannot be repaired by relaxing label equivalence.

\subsection{CLIP-Based Ablation: Diagnosing Taxonomy Mismatch vs.\ Semantic Confusion}

Table~\ref{tab:clip_ablation} analyzes IoU-matched detections whose labels disagree under closed-label scoring. We report the near-miss rate, defined as the fraction of mismatches that remain semantically close to the ground-truth label under the target vocabulary (text--text similarity above threshold and the ground-truth ranked in the top-$K$ candidates). We also summarize the strength of semantic proximity through the mean text similarity, the median ground-truth rank, and the top-5 inclusion rate.

Transfers that benefit from open-label evaluation exhibit substantially higher near-miss rates and stronger semantic proximity. For instance, Objects365$\rightarrow$BDD shows a 41\% near-miss rate with high mean similarity (0.73), a low median rank (2), and 88\% top-5 inclusion, indicating that many label ``errors'' are in fact semantically adjacent predictions. Objects365$\rightarrow$Cityscapes shows a similar pattern with 33\% near-misses and 81\% top-5 inclusion. By contrast, Cityscapes$\rightarrow$COCO has a near-miss rate of only 4\%, low similarity (0.58), and poor rank statistics, suggesting that its remaining errors are dominated by genuine semantic confusion and domain shift rather than taxonomy mismatch. 

\section{Qualitative Results}
\label{sec:qual}
Figure~4 presents qualitative results across different domain transfer settings. 
Figs.~4(b) and 4(f) demonstrate strong generalization, where predictions align closely with ground truth for Objects365$\rightarrow$Objects365 and Objects365$\rightarrow$Cityscapes transfers, indicating robust cross-domain feature learning. 
Similarly, Figs.~4(c), 4(d), and 4(n) show successful transfer to urban scenes, where major objects and pedestrians are localized accurately in Cityscapes-like environments, similarly, Figs.~4(q) and 4(s) illustrate strong cross-dataset transfer to daylight city scenes in BDD. However, performance degrades in challenging conditions such as nighttime (Fig.~4(t)) or glare (Fig.~4(r)) scenes. Although the Cityscapes-trained model is well attuned to urban layouts, its limited diversity compared to BDD100k prevents reliable generalization to these extreme cases. 
Figs.~4(a), 4(h), and 4(o) illustrate failure modes where overlapping objects or background clutter lead to incorrect predictions. 
Figs.~4(i) and 4(l) further highlight severe degradation under low-light conditions and large domain shifts, resulting in missed detections or misclassifications. While Fig.~4(g) achieves high IoU for Objects365$\rightarrow$COCO, it suffers from excessive hallucinated detections, reflecting a precision--recall trade-off. 
Finally, Figs.~4(j) and ~4(p) confirm that in-domain transfers (COCO$\rightarrow$COCO and Cityscapes$\rightarrow$Cityscapes) consistently yield the most reliable predictions.
\section{Discussion}
\label{sec: discussion}
\noindent \textbf{Setting-specific targets expose a robustness cliff.}
Transfer into driving datasets is extremely brittle: agnostic$\rightarrow$specific performance collapses in Table~\ref{tab:cross_dataset} and remains low even after open-label scoring in Table~\ref{tab:cross_dataset_openlabel}. This indicates that appearance, viewpoint, and context shift dominate over taxonomy mismatch in the hardest regimes.

\noindent \textbf{Cross-dataset shift is directional.}
Many pairs are strongly asymmetric in Table~\ref{tab:cross_dataset}, so ``domain gap'' cannot be treated as symmetric. Reporting one-way curated source$\rightarrow$target results can hide the harder direction and overstate robustness.

\noindent \textbf{Label mismatch is real but bounded, and diagnostics reveal its structure.}
Open-label evaluation yields consistent yet limited gains (Table~\ref{tab:cross_dataset_openlabel}), while near-miss statistics (Table~\ref{tab:clip_ablation}) distinguish semantically close mismatches from genuine semantic confusion. Large residual drops after open-label scoring point to robustness failures beyond label alignment.




\begin{figure}[H]
\centering

\begin{subfigure}{0.23\linewidth}
    \includegraphics[width=\linewidth]{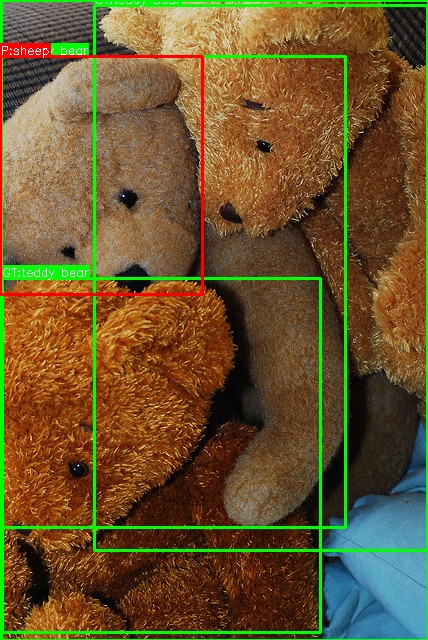}
    \caption{O365$\rightarrow$COCO}
\end{subfigure}\hfill
\begin{subfigure}{0.23\linewidth}
    \includegraphics[width=\linewidth]{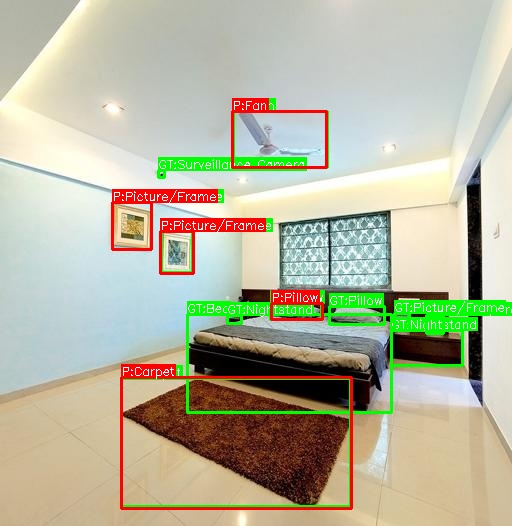}
    \caption{O365$\rightarrow$O365}
\end{subfigure}\hfill
\begin{subfigure}{0.23\linewidth}
    \includegraphics[width=\linewidth]{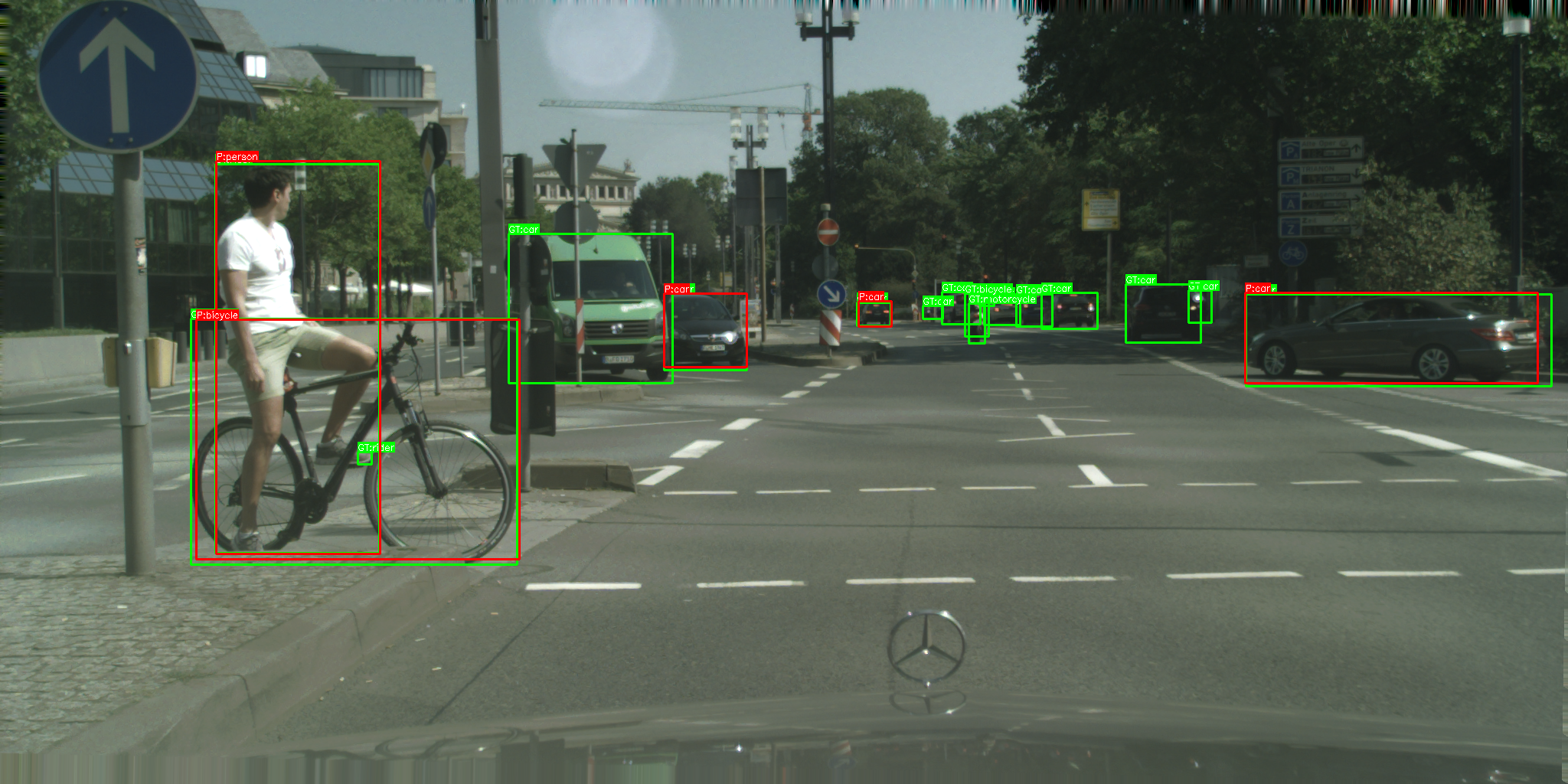}
    \caption{O365$\rightarrow$City}
\end{subfigure}\hfill
\begin{subfigure}{0.23\linewidth}
    \includegraphics[width=\linewidth]{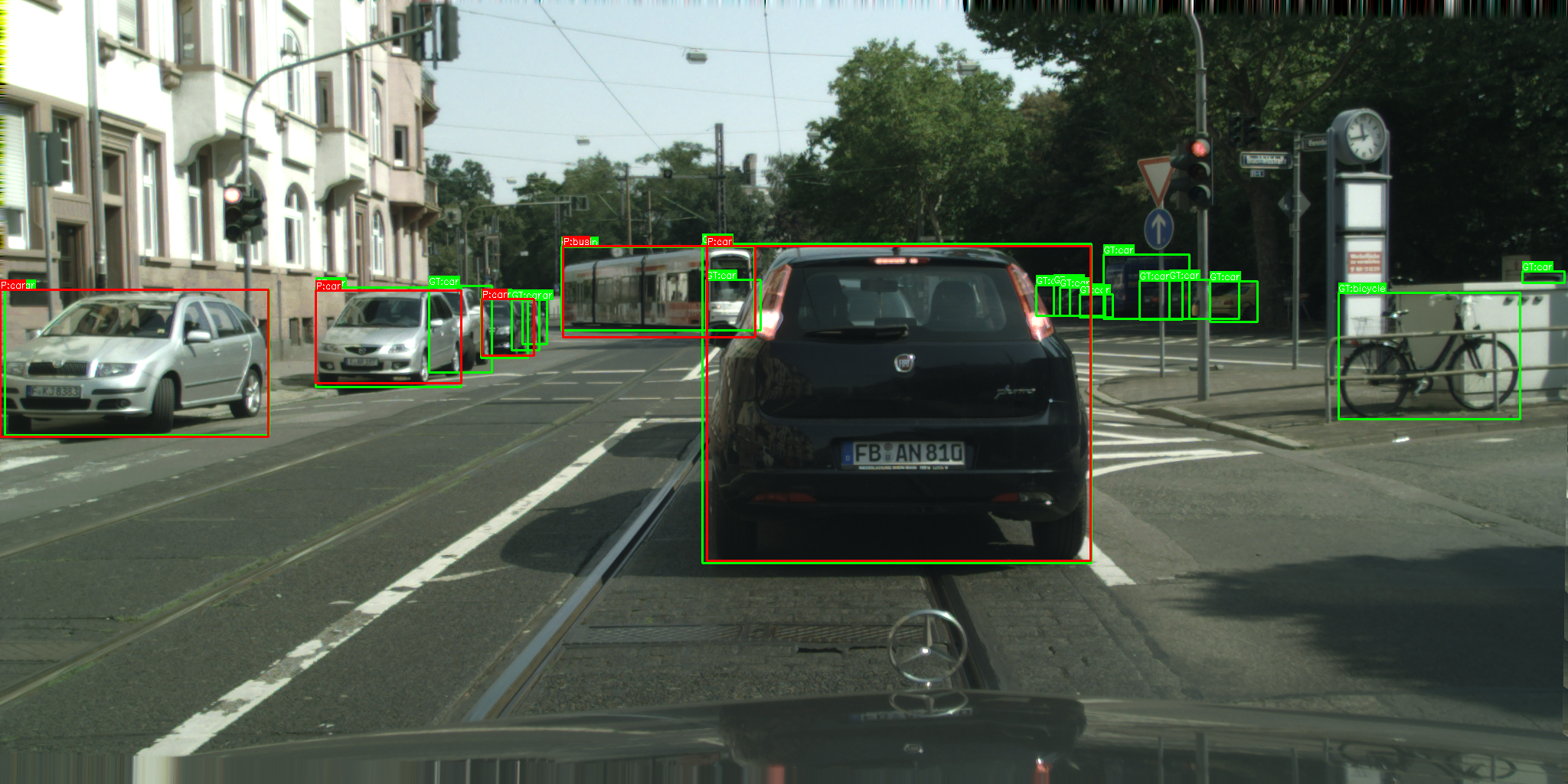}
    \caption{COCO$\rightarrow$City}
\end{subfigure}

\vspace{0.25cm}

\begin{subfigure}{0.23\linewidth}
    \includegraphics[width=\linewidth]{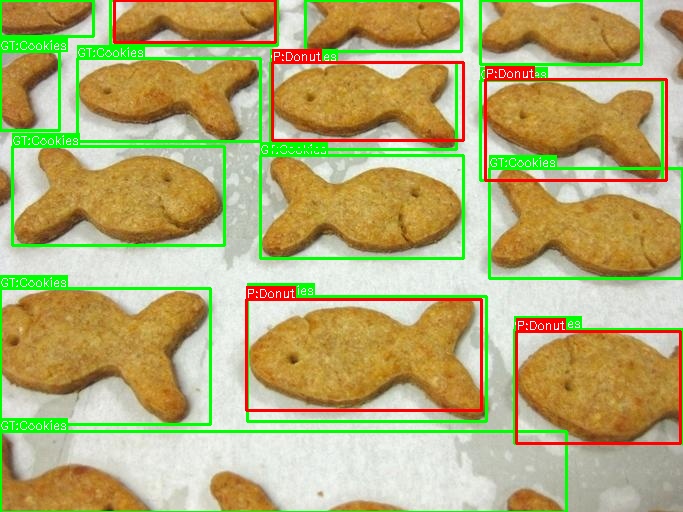}
    \caption{COCO$\rightarrow$O365}
\end{subfigure}\hfill
\begin{subfigure}{0.23\linewidth}
    \includegraphics[width=\linewidth]{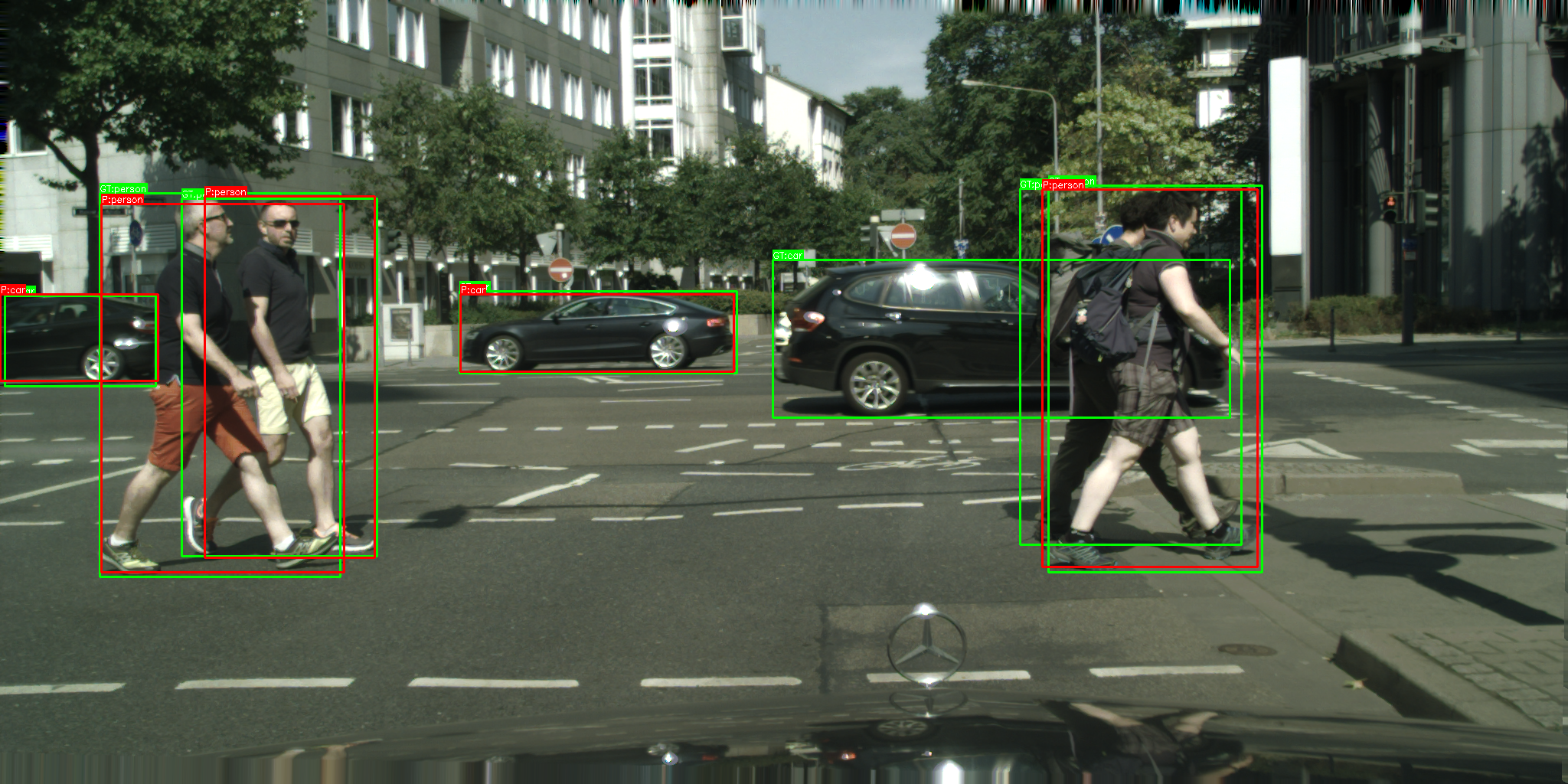}
    \caption{O365$\rightarrow$City}
\end{subfigure}\hfill
\begin{subfigure}{0.23\linewidth}
    \includegraphics[width=\linewidth]{best_2_iou_0.964_000000000285.jpg}
    \caption{O365$\rightarrow$COCO}
\end{subfigure}\hfill
\begin{subfigure}{0.23\linewidth}
    \includegraphics[width=\linewidth]{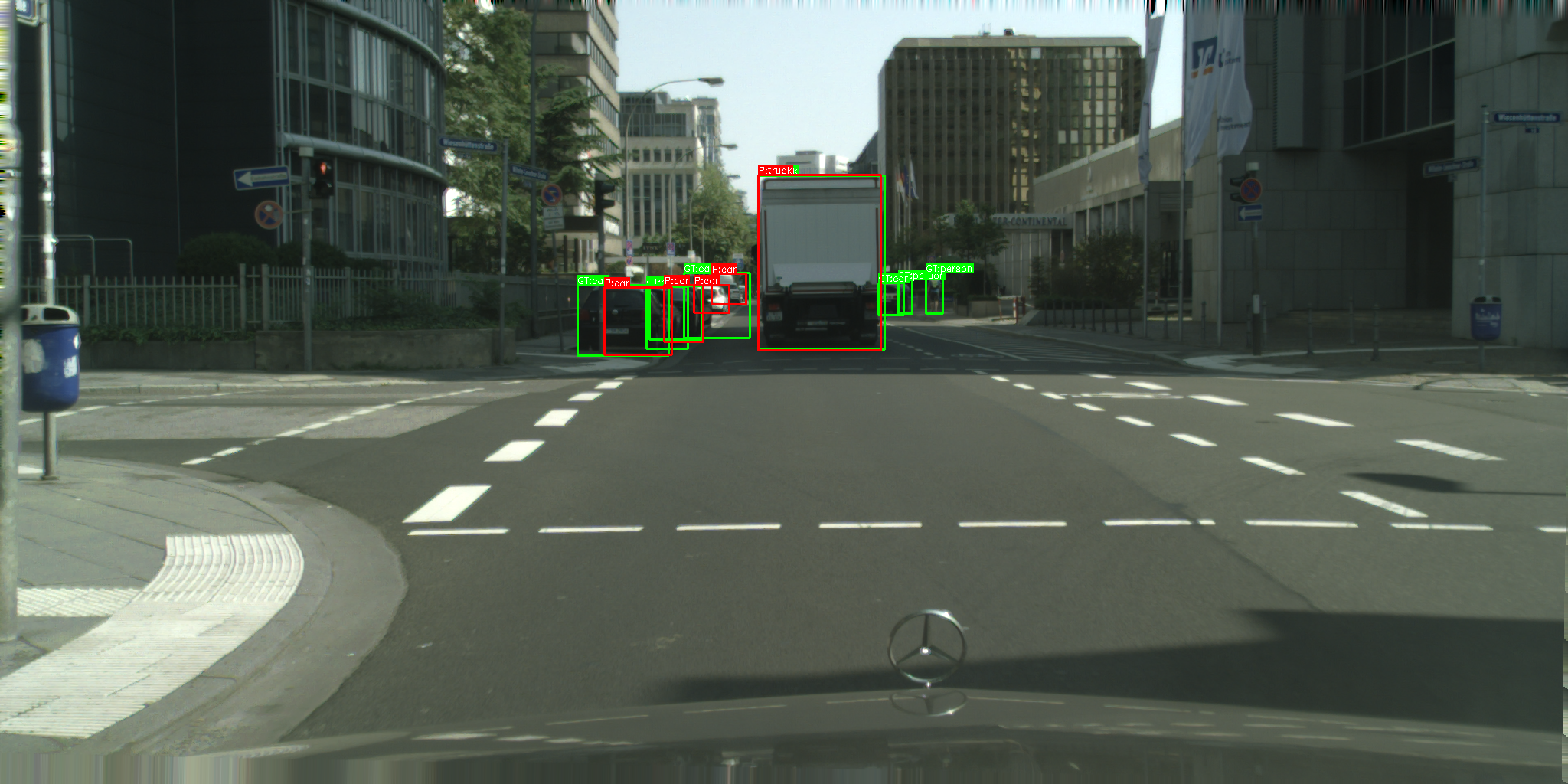}
    \caption{COCO$\rightarrow$City}
\end{subfigure}

\vspace{0.25cm}

\begin{subfigure}{0.23\linewidth}
    \includegraphics[width=\linewidth]{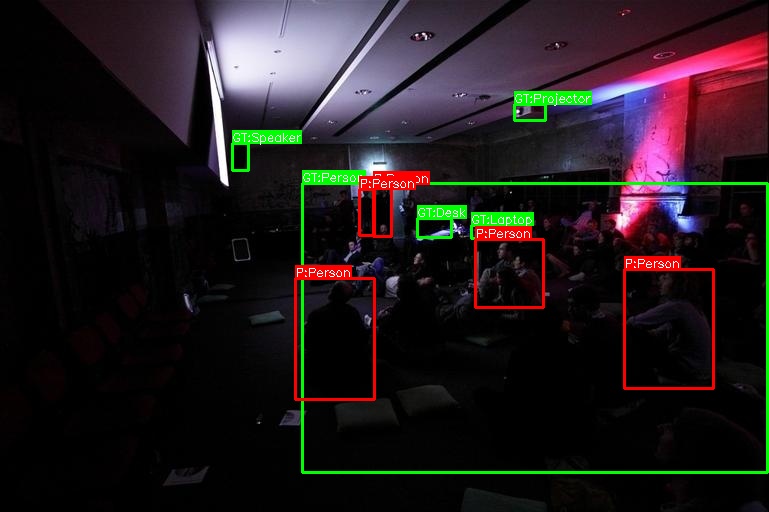}
    \caption{COCO$\rightarrow$O365}
\end{subfigure}\hfill
\begin{subfigure}{0.23\linewidth}
    \includegraphics[width=\linewidth]{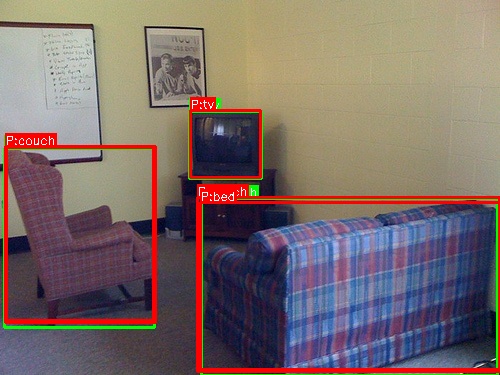}
    \caption{COCO$\rightarrow$COCO}
\end{subfigure}\hfill
\begin{subfigure}{0.23\linewidth}
    \includegraphics[width=\linewidth]{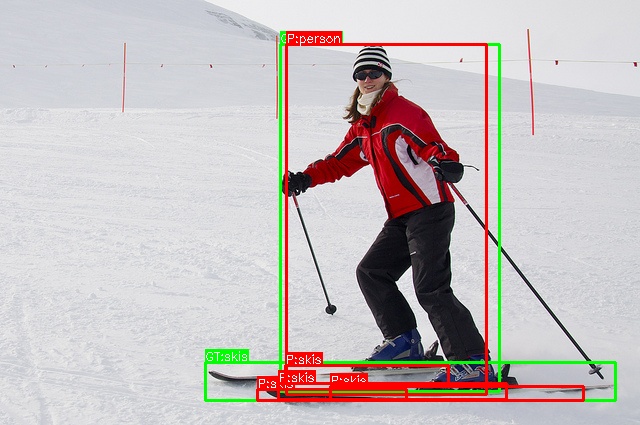}
    \caption{COCO$\rightarrow$COCO}
\end{subfigure}\hfill
\begin{subfigure}{0.23\linewidth}
    \includegraphics[width=\linewidth]{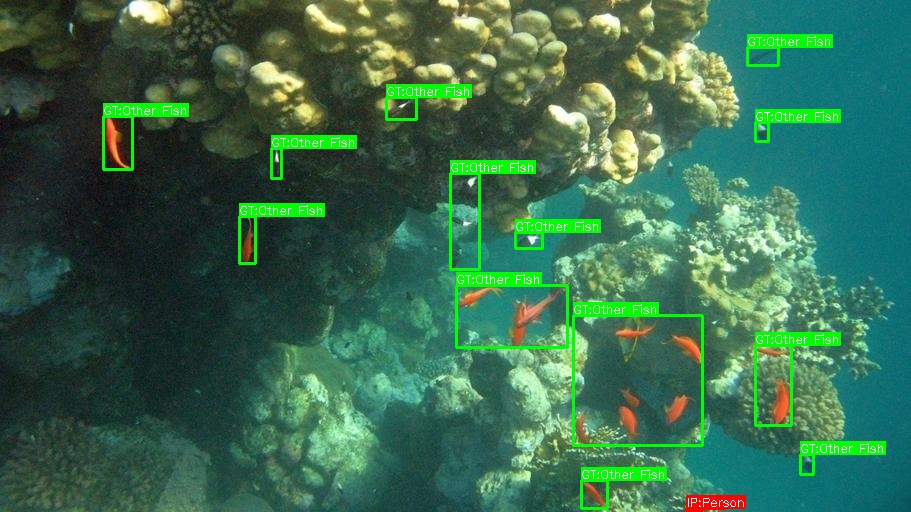}
    \caption{City$\rightarrow$O365}
\end{subfigure}

\vspace{0.25cm}

\begin{subfigure}{0.23\linewidth}
    \includegraphics[width=\linewidth]{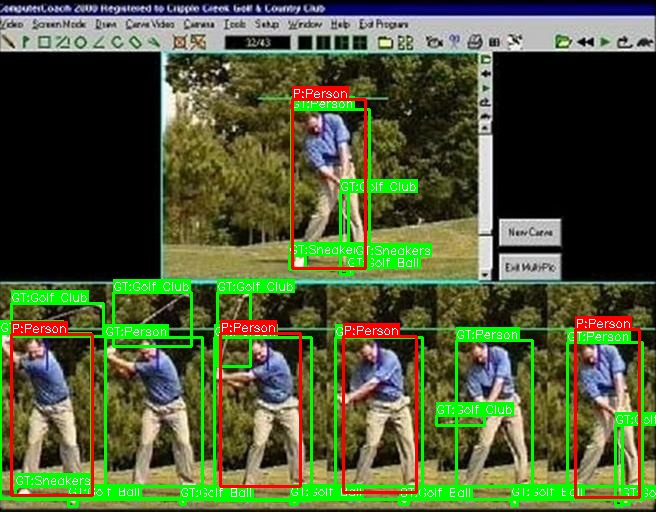}
    \caption{City$\rightarrow$O365}
\end{subfigure}\hfill
\begin{subfigure}{0.23\linewidth}
    \includegraphics[width=\linewidth]{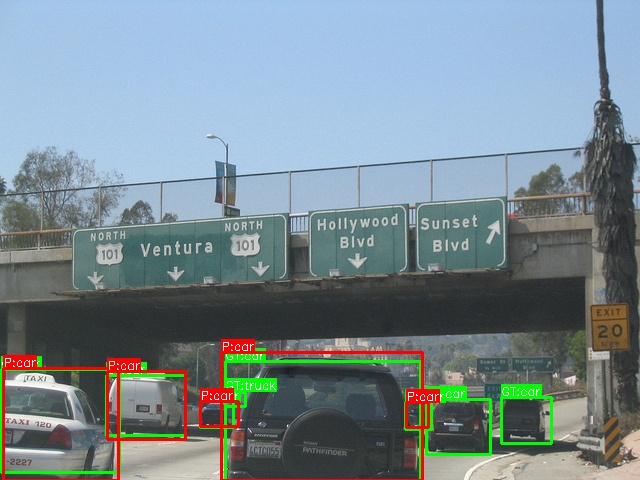}
    \caption{City$\rightarrow$COCO}
\end{subfigure}\hfill
\begin{subfigure}{0.23\linewidth}
    \includegraphics[width=\linewidth]{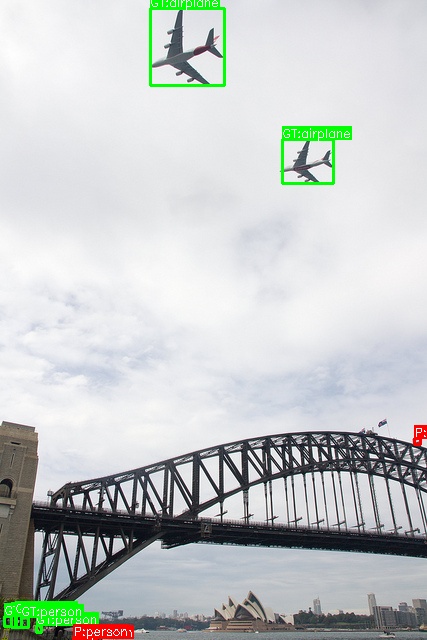}
    \caption{City$\rightarrow$COCO}
\end{subfigure}\hfill
\begin{subfigure}{0.23\linewidth}
    \includegraphics[width=\linewidth]{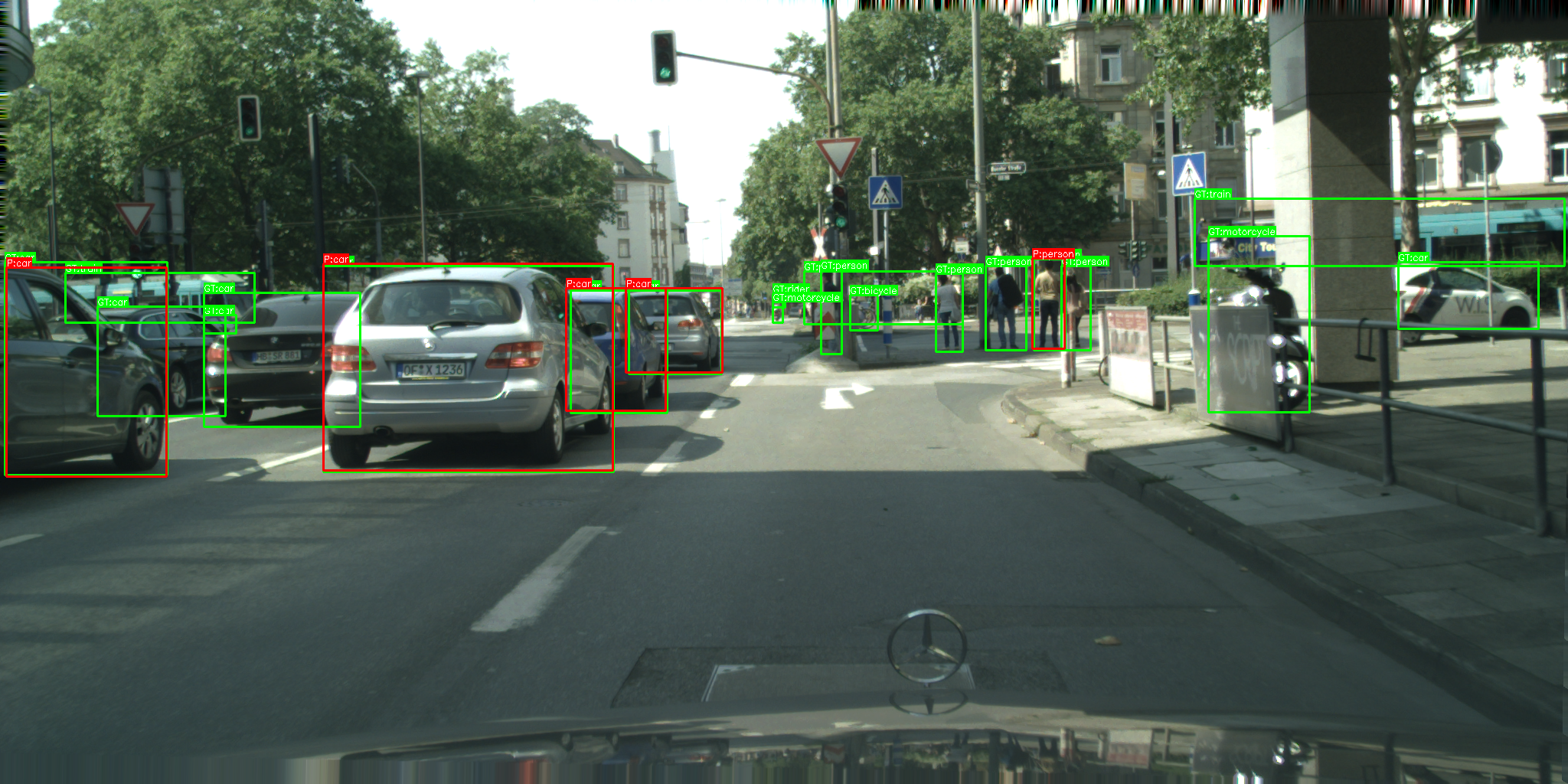}
    \caption{City$\rightarrow$City}
\end{subfigure}

\begin{subfigure}{0.23\linewidth}
    \includegraphics[width=\linewidth]{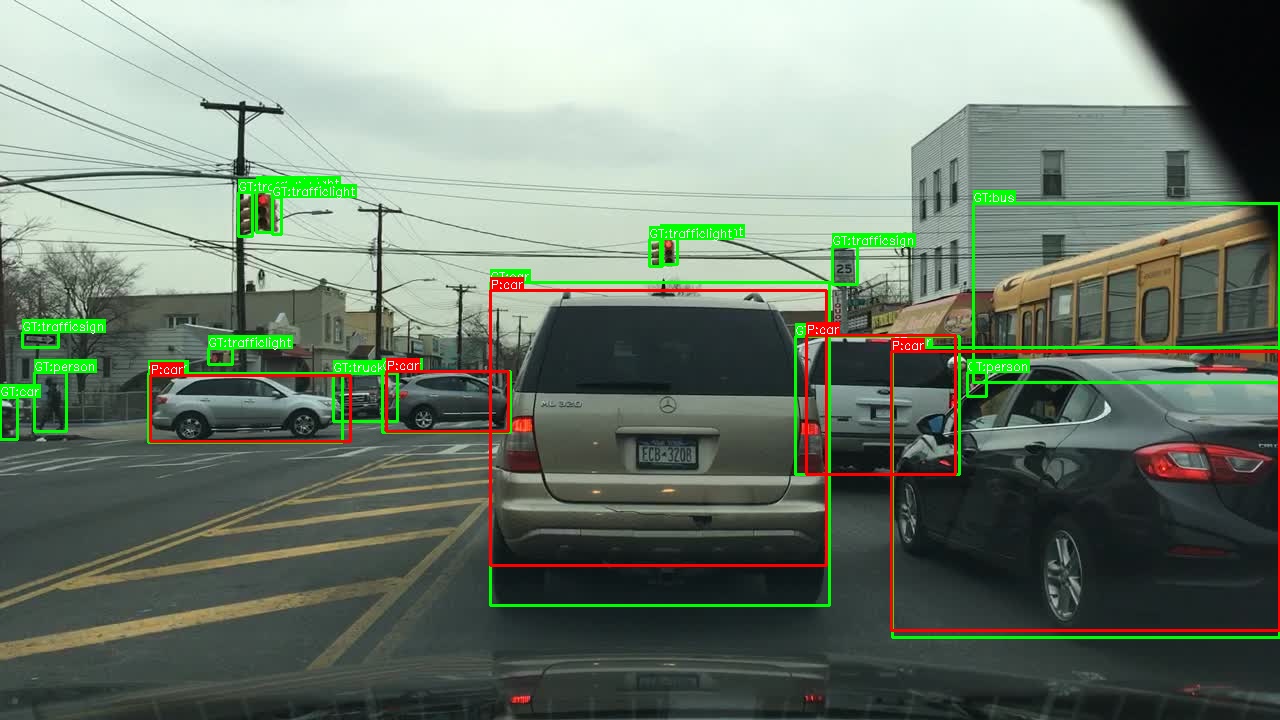}
    \caption{COCO$\rightarrow$BDD}
\end{subfigure}\hfill
\begin{subfigure}{0.23\linewidth}
    \includegraphics[width=\linewidth]{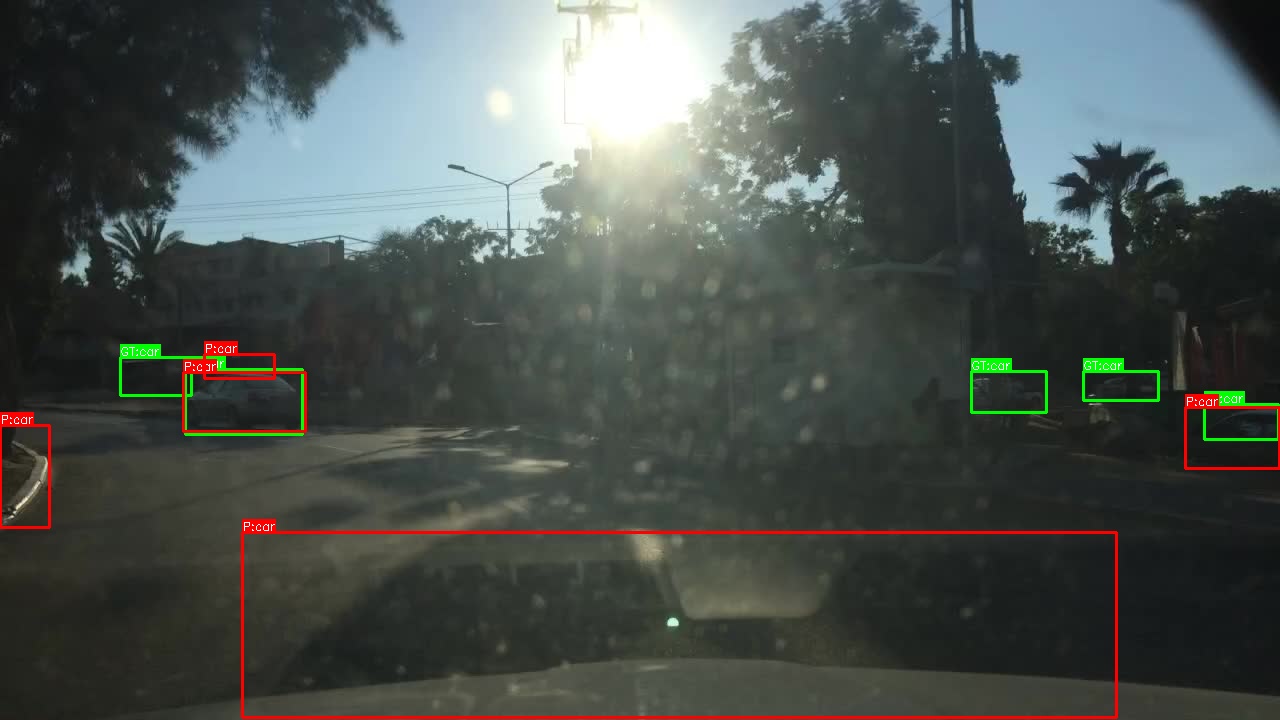}
    \caption{COCO$\rightarrow$BDD}
\end{subfigure}\hfill
\begin{subfigure}{0.23\linewidth}
    \includegraphics[width=\linewidth]{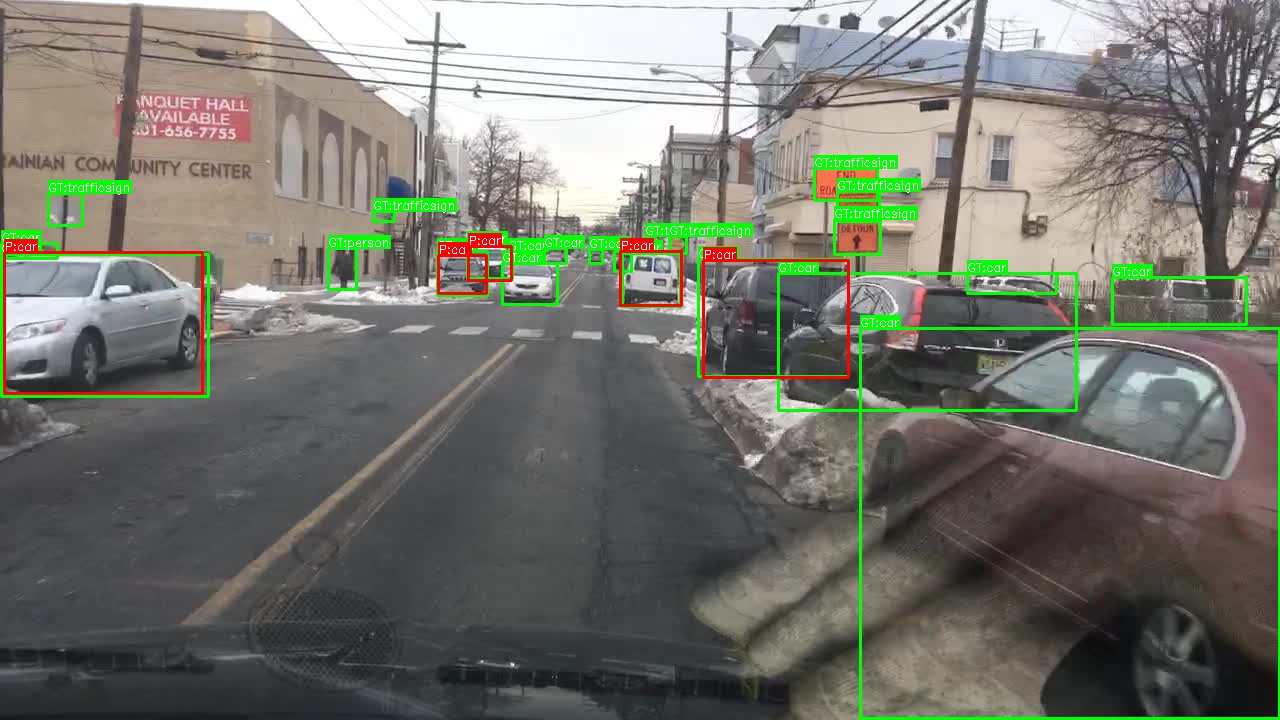}
    \caption{City$\rightarrow$BDD}
\end{subfigure}\hfill
\begin{subfigure}{0.23\linewidth}
    \includegraphics[width=\linewidth]{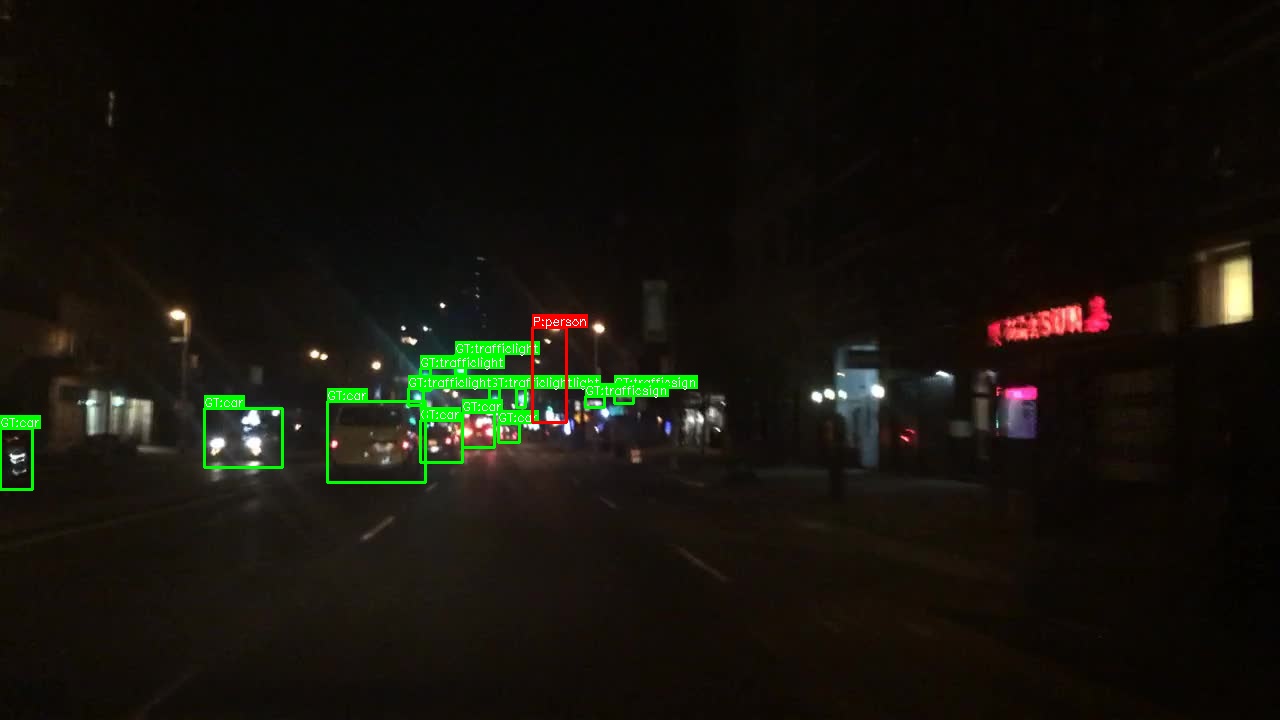}
    \caption{City$\rightarrow$BDD}
\end{subfigure}

\vspace{0.25cm}
\caption{In-domain and Cross-domain evaluation results. City refers to images from Cityscapes, and O365 refers to images from the Objects365 dataset. X$\rightarrow$Y refers to the inference setting of the model being trained on the dataset X and tested on the dataset Y}
\label{fig:results}
\end{figure}

\section{Conclusion}

\label{sec:conclusion}
In this work, we study cross-dataset object detection under a setting-specificity-based framework. We organized benchmarks into setting-agnostic and setting-specific categories, and evaluated a standard detection model on them across all train-test pairs, running inference under both closed-label and open-label settings in order to explicitly isolate semantic misalignment from visual domain shift. Notably, the most pronounced performance degradations occur when transferring from diverse, setting-agnostic datasets to narrowly defined, setting-specific environments, underscoring the dominant influence of domain shift even after addressing label mismatches. Our analysis has implications for safety-critical deployments, where detectors trained on standard curated datasets must operate on unseen conditions. Our study is limited to only standard detector architectures, and a fixed set of benchmarks, which may not capture all forms of real-world distribution shifts. Future work can explore domain-adaptation strategies to overcome these shifts, to improve robustness, while further investigating the interplay between setting-agnostic and setting-specific data.

\bibliographystyle{splncs04}
\bibliography{ICPR_2026_LaTeX_Templates/samplepaper}




\end{document}